# Trading off Completeness for Efficiency — The PARSETALK Performance Grammar Approach to Real-World Text Parsing

Peter Neuhaus     Udo Hahn

Freiburg University
Computational Linguistics Lab
Europaplatz 1, D-79085 Freiburg, Germany
{neuhaus,hahn}@coling.uni-freiburg.de

**Abstract**

We argue for a performance-based design of natural language grammars and their associated parsers in order to meet the constraints posed by real-world natural language understanding. This approach incorporates declarative and procedural knowledge about language and language use within an object-oriented specification framework. We discuss several message passing protocols for real-world text parsing and provide reasons for sacrificing completeness of the parse in favor of efficiency.

## 1 INTRODUCTION

Over the past decades the design of natural language grammars and their parsers was almost entirely based on *competence* considerations (Chomsky, 1965). These hailed pure declarativism (Shieber, 1986) and banned procedural aspects of natural language use out of the domain of language theory proper. The major premises of that approach were to consider sentences as the primary linguistic object of investigation, to focus on syntactic descriptions, and to rely upon perfectly well-formed utterances for which complete grammar specifications of arbitrary depth and sophistication were available. In fact, promising efficiency results can be achieved for parsers under corresponding optimal laboratory conditions. Considering, however, the requirements of natural language *understanding*, i.e., the integration of syntax, semantics, and pragmatics, and taking *ill-formed* input or *incomplete* knowledge into consideration, the processing costs either tend to increase at excessive rates or linguistic processing even fails completely.

As a consequence, the challenge to meet the specific requirements imposed by real-world texts has led many researchers in the NLP community to re-engineer competence grammars and their parsers and to provide various add-ons in terms of constraints (Uszkoreit, 1991), heuristics (Huyck & Lytinen, 1993), statistics-based weights (Charniak, 1993), etc. In contradistinction to these approaches, our principal goal has been to incorporate performance conditions already in the design of natural language grammars, yielding so-called *performance grammars*. Thus, not only declarative knowledge (as is common for competence grammars), but also *procedural* knowledge (about control and parsing strategies, resource limitations, etc.) has to be taken into consideration at the *grammar specification* level proper. This is achieved by providing self-contained description primitives for the expression of procedural knowledge. We have taken considerable care to transparently separate declarative (structure-oriented) from procedural (process-oriented) knowledge pieces. Hence, we have chosen a formally homogeneous, highly modularized object-oriented grammar specification framework, namely the actor model of computation which is based on concurrently active objects that communicate by asynchronous message passing (Agha, 1990).

The PARSETALK system whose design is based on these performance considerations (Hahn et al., 1994) forms part of a text knowledge acquisition system whose input comes from two domains, viz. test reports from the information technology field and medical findings reports. The analysis of texts (instead of isolated sentences) requires, first of all, the consideration of textual phenomena by a dedicated *text grammar* (local ones such as anaphora (Strube & Hahn, 1995), or even global coherence relations (Hahn, 1992)). Second, text understanding is based on the execution of inferences by which text propositions are integrated into the text knowledge base with reference to a canonical representation of the *background knowledge* of the underlying domain. This way, grammatical (language-specific) and conceptual (domain-specific) knowledge are closely coupled. Third, text understanding in humans occurs immediately and at least within specific processing cycles in parallel (Thibadeau et al., 1982). These processing strategies we find in human language processing are taken as hints how the complexity of natural language understanding can reasonably be overcome by machines. Thus, text parsing devices should operate *incrementally*

cmp-lg/9605026  15 May 1996

and *concurrently*. In addition, the consideration of *real-world* texts (instead of researcher-generated material covered by complete specifications) forces us to supply mechanisms which allow for the *robust* processing of extra- and ungrammatical input. We take an approach where — in the light of abundant specification gaps at the grammar and domain representation level — the degree of underspecification of the knowledge sources or the impact of grammar violations directly corresponds to a lessening of the precision and depth of text knowledge representations, thus aiming at a sophisticated *fail-soft* model of *partial* text parsing.

## 2   THE PARSETALK GRAMMAR

The PARSETALK performance grammar contains fully *lexicalized* grammar specifications in terms of configurational constraints on word classes and morphological features as well as on word order and conceptual compatibility. Grammatical conditions of these types are combined in terms of *valency* constraints (at the phrasal and clausal level) as well as *textuality* constraints (at the text level of consideration), which concrete dependency structures and local as well as global coherence relations must satisfy. The compatibility of grammatical features (encapsulated by methods we refer to as SYNTAXCHECK) is computed by a unification mechanism, while the evaluation of semantic and conceptual constraints (we here refer to as CONCEPTCHECK) relies upon terminological reasoning in terms of subsumption. Thus, while the dependency relations represent the linguistic structures of the input, the conceptual relations yield the targeted representation of the text content (for an illustration, cf. Fig. 7).

In order to structure the underlying lexicon, *inheritance* mechanisms are used. Lexical specifications are thus organized along the grammar hierarchy at various abstraction levels, e.g., with respect to generalizations on word classes. Lexicalization of this form already yields a fine decomposition of declarative grammar knowledge. It lacks, however, an appropriate description equivalent at the procedural level. We therefore provide lexicalized communication primitives to allow for heterogeneous and local forms of interaction among lexical items.

Semantic and domain knowledge specifications are based on a common hybrid terminological, classification-based knowledge representation language (for a survey, cf. Woods & Schmolze (1992)). Ambiguities which result in interpretation variants are managed by a context mechanism of the underlying KB system (there is no distinction at the representational level between semantic and conceptual interpretations of texts).

*Robustness* at the grammar level is achieved by several means. Dependency grammars describe binary, functional relations between words rather than contiguous constituent structures. Thus, ill-formed input often does still have an (incomplete) analysis in our grammar. Furthermore, it is possible to specify words at different levels of syntactic or semantic granularity. The main burden of robustness, however, is assigned to a dedicated message passing protocol discussed in the next section.

## 3   THE PARSETALK PARSER

At the class level of object-oriented grammar specifications, we represent lexical items as *word actors*, which, when instantiated, are acquainted with other actors representing the head or modifiers in the current utterance.

A specialized actor type, the *phrase actor*, groups word actors which are connected by dependency relations and encapsulates administrative information about that phrase. Accordingly, a message does not have to be sent directly to a specific word actor, but will be sent to the mediating phrase actor which forwards it to an appropriate word actor. Furthermore, the phrase actor holds the communication channel to the corresponding interpretation context in the domain knowledge base system.

A *container actor* encapsulates several phrase actors that constitute alternative analyses for the *same* part of the input text (i.e., structural ambiguities). Container actors play a central role in controlling the parsing process, because they keep information about the *textually* related (*preceding*) container actors holding the left context and the *chronologically* related (*previous*) container actors holding a part of the head-oriented parse history.

**Basic Parsing Protocol.**  We use a graphical description language to sketch the message passing protocol for establishing dependency relations as depicted in Fig. 1 (the phrase actor's active head is visualized by ⊕). A searchHeadFor message (and *vice versa* a searchModifierFor message if searchHeadFor fails to succeed) is sent to the textually preceding container actor (precedence relations are depicted by bold dashed lines), which simultaneously directs this message to its encapsulated phrase actors. At the level of a single phrase actor, the distribution of the searchHeadFor message occurs for all word actors at the "right rim" of the dependency tree (depicted by ⊙). A word actor that receives a searchHeadFor message from another word actor tests whether a dependency relation can be established by checking its corresponding valency constraints (applying SYNTAXCHECK and CONCEPTCHECK). In case of successful evaluation, a headFound message is returned, the sender and the receiver are copied (to enable alternative attachments in the concurrent system, i.e., no destructive operations are carried out), and a dependency relation, indicated by a dotted line, is established between those copies which join into a phrasal relationship. For illustration purposes, consider the analysis of a phrase like *"Zenon sells this printer"* covering the content of the phrase actor which textually precedes the phrase actor holding the dependency structure for *"for $2,000"*. The latter actor requests its attachment as a modifier to some head. The resultant new container actor (encapsulating the dependency analysis for *"Zenon sells this printer for $2,000"* in two phrase actors) is, at the same time, the historical successor to the phrase actor covering the analysis for *"Zenon sells this printer"*.

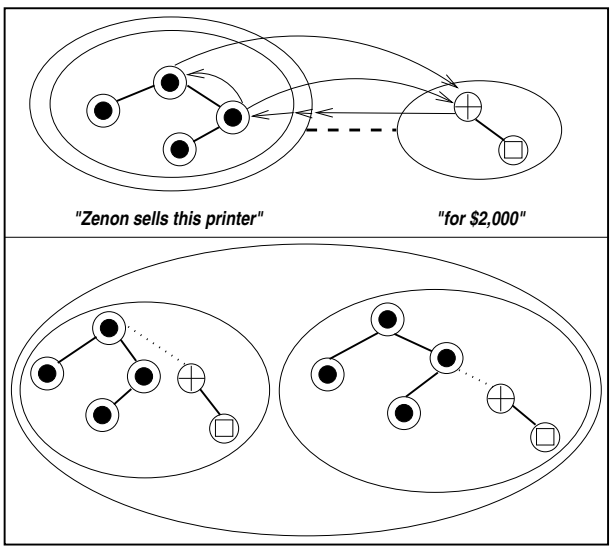

Figure 1: Basic Mode (incl. Structural Ambiguities)

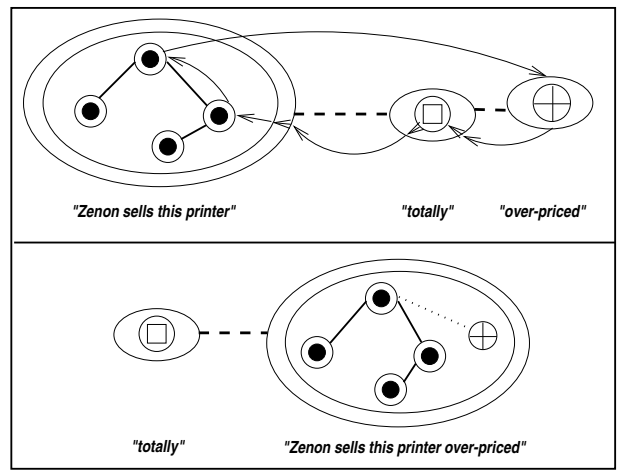

Figure 2: Skipping Mode

The structural ambiguity inherent in the example is easily accounted for by this scheme. The criterion for a structural ambiguity to emerge is the reception of at least two positive replies to a single searchHeadFor (or searchModifierFor) message by the initiator (cf. Fig. 1). The basic protocol already provides for the concurrent copying and feature updates required. In the example from Fig. 1, two alternative readings are parsed, one phrase actor holding the attachment to the verb (*"sells"*), the other holding that to the noun (*"printer"*). The crucial point about these ambiguous syntactic structures is that they have conceptually different representations in the domain knowledge base. In the case of Fig. 1 verb attachment leads to the instantiation of the PRICE slot of the corresponding SELL action, while the noun attachment leads to the corresponding instantiation of the PRICE slot of PRINTER.

**Robustness: Skipping Protocol.** The principled lack of exhaustive linguistic and conceptual specifications requires mechanisms for a fail-soft parsing behavior. An example for this is shown in Fig. 2 which illustrates a typical "skipping" scenario. The currently active container addresses a searchHeadFor (or searchModifierFor) message to its textually immediately preceding container actor. If *both* types of messages fail, the immediately preceding container of the active container forwards these messages — in the canonical order — to its immediately preceding container actor. If any of these two message types succeeds after that mediation, a corresponding (discontinuous) dependency structure is built up. Furthermore, the skipped container is moved to the left of the newly built container actor. Note that this behavior results in the reordering of the lexical items analyzed so far in terms of textual precedence such that skipped containers are continuously moved to the left. As an example, consider the phrase *"Zenon sells this printer"* and let us further assume *"totally"* to be a grammatically unknown item which is followed by the occurrence of *"over-priced"* as the active container. Skipping yields a structural analysis for *"Zenon sells this printer over-priced"*, while *"totally"* is simply discarded from further consideration. This mode requires an extension of the basic protocol in that searchHeadFor and searchModifierFor messages are forwarded across non-contiguous parts of the analysis when these messages do not yield a positive result for the requesting actor relative to the *immediately* adjacent container actor.

**Backtracking Protocol.** The backtracking mode characterizes a state of the structural analysis where none of the aforementioned protocols terminate successfully in *any* textually preceding container, i.e., several repeated skippings have occurred, until a linguistically plausible barrier is encountered. In this case, backtracking takes place and messages are now directed to *historically* previous containers, i.e., to containers holding the parse history. This is realized in terms of a protocol extension by which searchHeadFor (or searchModifierFor) messages, first, are reissued to the *textually* immediately preceding container actor which then forwards these messages to its *historically* previous container actor. This actor contains the head-centered results of the analysis of the left context prior to the structural extension held by the historical successor.[1] Relation attachments for heads or modifiers are now checked between the historically preceding container and the active container as depicted in Fig. 3.

---

[1] Any container which holds the modifying part of the structural analysis of the historical successor is deleted. Hence, the deletion of

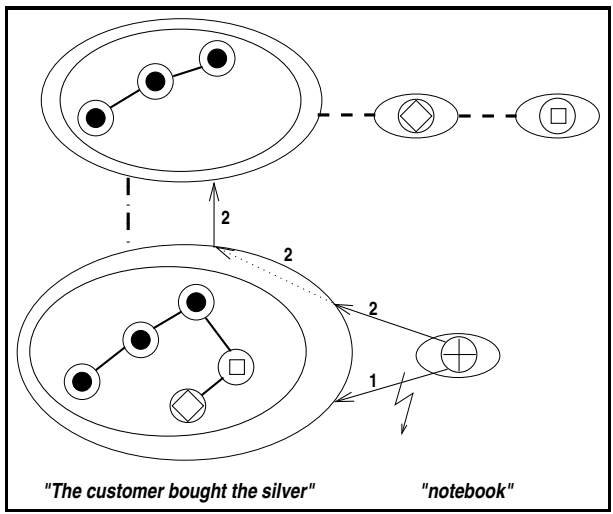

Figure 3: Backtracking Mode

If the valency constraints are met, a composite phrase actor is formed (cf. Fig. 4) with a necessarily discontinuous analysis. A slightly modified protocol allows the skipped items to try a reanalysis such that the reSearch-HeadFor (or reSearchModifierFor) messages they send to the new container actor are directly forwarded to those word actors in the composite phrase actor where the discontinuity occurs. As an example, consider the fragment *"the customer bought the silver"* (with *"silver"* holding the noun reading, cf. Fig. 3). This yields a perfect analysis the result of which, however, cannot be further augmented when the word actor *"notebook"* asks next for a possible attachment.[2] Two intervening steps of reanalysis yield the final structural configuration depicted in Fig. 4.

**Preference Encoding.** In the model we have sketched so far, messages are directed from the active container actor to its textually preceding container actor and are sent *in parallel* to *all* the phrase actors encapsulated by that container actor. However, the management of ambiguities can be made more efficient by structuring the set of ambiguous analyses in that preference is given to a particular subset of these phrase actors. A predicate is supplied that separates a set of possible attachments into preferred and deferred ones. Then the former ones can be collected into the active container, while the latter ones are stored in a separate container. Thus, they are not pursued in the

---

containers not contained in historically previous head-centered parses, obviously, may result in the parser becoming incomplete in spite of backtracking.

[2] As PARSETALK is an arc-eager parsing system, a possible dependency relation will always be established and favored over an alternative structural description that cannot be immediately attached. Hence, the adjective reading of *"silver"* will not be considered in the initial analysis.

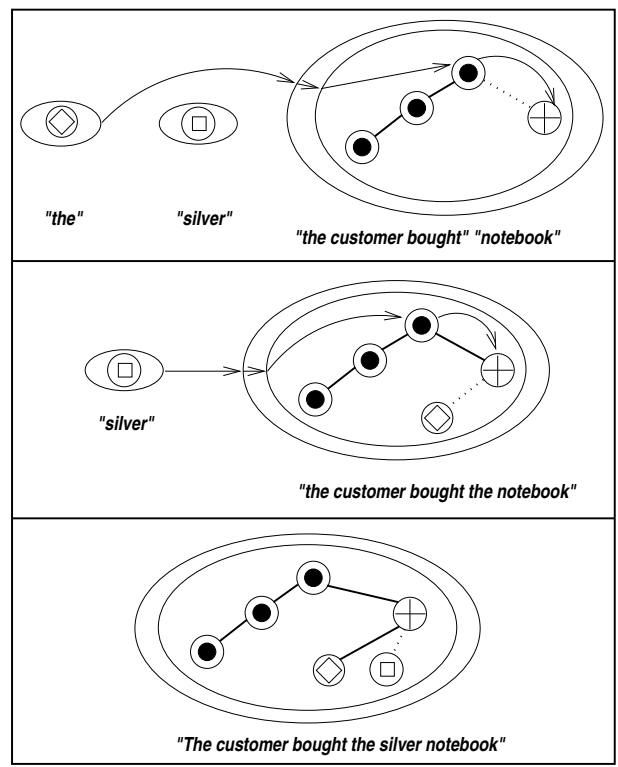

Figure 4: Backtracking Mode (cont.)

current analysis but are retained for backtracking.

**Prediction Protocol.** The depth-first approach of the parser brings about a decision problem whenever a phrase cannot be integrated into (one of) the left context analyses. The reason for this indeterminacy is twofold: Either, the left context and the current phrase are to be related by a word not yet read from the input and, thus, the analysis should proceed without an attachment. Or, depth-first analysis was misguided and a backtrack should be invoked to revise a former decision with respect to attachment information available by now.

We follow a prediction approach, where certain words predict the *word class* of their head and/or mandatory modifiers. In addition to the basic searchHeadFor and searchModifierFor protocols, predicted word actors (cf. Fig. 5) process a searchPredictionFor protocol by which a word actor is unified with its predicted "placeholder" if its class is equal or more specific. To attach a modifier to a predicted word we exploit the class hierarchy and add any contextually plausible valency of a *subclass* to the valency frame of the predicted word. Under these conditions, the analysis can proceed more informed.

**Text Phenomena.** A particularly interesting feature of the PARSETALK grammar is its capability to seamlessly integrate the sentence and text level of linguistic analysis. The protocol which accounts for local text coherence analysis makes use of a special actor, the *centering actor*, which keeps a backward-looking center ($C_b$) and a preferentially ordered list of forward-looking centers ($C_f$) of the previous utterance (we here adapt the well-known centering model by Grosz et al. (1995)). These lists are accessed to establish proper referential links between an anaphoric expression in the current utterance and the valid antecedent in preceding ones. Nominal anaphora (cf. the occurrences of *"the company"* and *"these printers"* in Fig. 6) trigger a special searchNomAntecedent message. When it reaches the $C_f$ list, possible antecedents are accessed in the given preference order. If an antecedent and the anaphor fulfil certain grammatical and conceptual constraints (Strube & Hahn, 1995) an antecedentFound message is issued to the anaphor, and finally, the discourse referent of the antecedent replaces the one in the original anaphoric expression in order to establish local coherence. The effects of these changes at the level of text knowledge structures are depicted in Fig. 7, which contains the terminological representation structures for the sentences in Fig. 6.

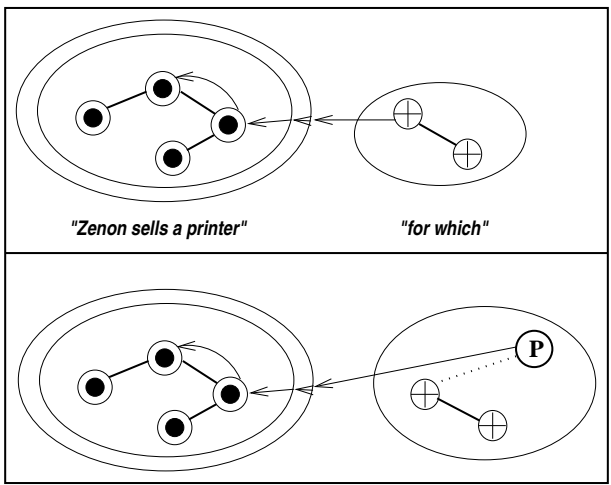

Figure 5: Prediction Mode

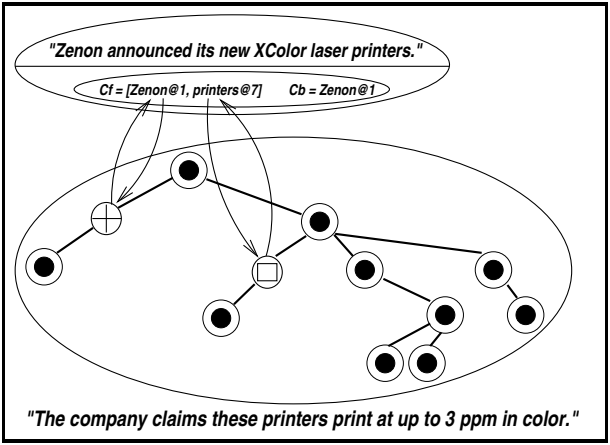

Figure 6: Anaphora Resolution Mode

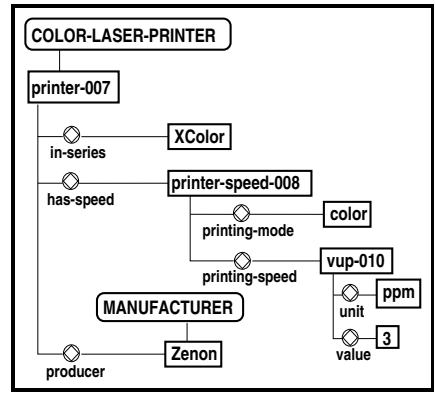

Figure 7: Sample Output of Text Parsing

## 4 COMPLETENESS *vs.* EFFICIENCY

The requirements of real-world text understanding pose severe complexity problems for parsers in such an operational environment. A depth-first parsing strategy usually will increase the parsing efficiency in the *average case*, i.e., for inputs that are in accordance with the parser's preferences. For the rest of the inputs backtracking is necessary to get the correct analyses. Thus, the *worst case* for a depth-first strategy applies to input which cannot be assigned any analysis at all (i.e., in cases of extra- or ungrammaticality). Such a failure situation leads to an exhaustive backtracking in a computationally complete parsing algorithm such that breadth-first and depth-first complexities will coincide.

With respect to real-world text inputs we have to face such worst-cases in many sentences. That is, by a complete depth-first strategy we merely trade space for time complexity. To maintain the gained efficiency of depth-first operation it is necessary to prevent the parser from exhaustive backtracking. In the PARSETALK parser this is achieved by two means. First, by restricting memoization of attachment candidates for backtracking (e.g., by retaining only the head portion of a newly built phrase, cf. footnote 1). Second, by restricting the accessibility of attachment candidates for backtracking (e.g., by bounding the forwarding of backtracking messages to linguistically plausible barriers such as punctuation actors). In effect, these restrictions render the parser computationally incomplete, since some inputs correctly covered by the grammar specification cannot be analyzed by the parser. Thus, the trade-off between robustness, efficiency, and completeness becomes obvious.

The efficiency gain that results from deciding against completeness is empirically demonstrated by a comparison of the PARSETALK system with a standard active chart parser[3]. Since the chart parser is implemented in

---
[3]The chart parser (Winograd, 1983) was adapted to parsing a dependency grammar. No packing or structure sharing techniques could be used, since the analyses are continuously interpreted in conceptual terms.

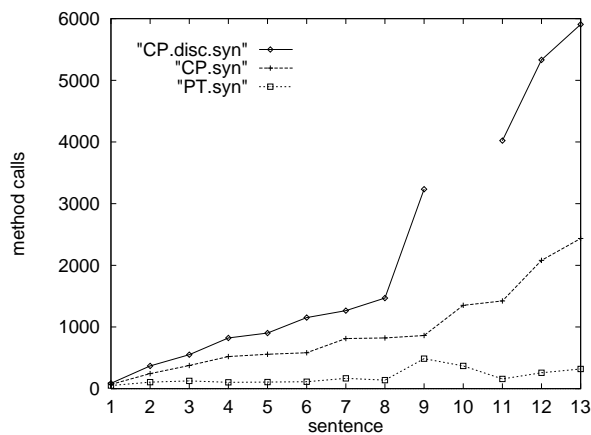

Figure 8: Calls to SYNTAXCHECK

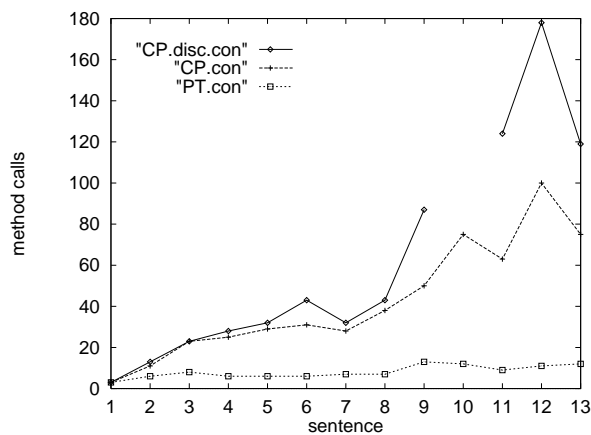

Figure 9: Calls to CONCEPTCHECK

Smalltalk, while the PARSETALK system is implemented in Actalk (Briot, 1989) — a language which simulates the concurrent execution of actors —, a direct comparison of run times is not meaningful (though even at that level of consideration the PARSETALK system outperforms the chart parser). We rather compare the number of method executions given exactly the same dependency grammar. The computationally most expensive methods we consider are SYNTAXCHECK and CONCEPTCHECK (cf. Section 2). Especially the latter consumes large computational resources, since for each interpretation variant a context has to be built in the KB system and conceptual consistency must be checked. The number of calls to these methods for a sample of 13 increasingly complex sentences from the information technology domain test library is given by the plots in Figs. 8 ("CP.syn" and "PT.syn", respectively) and 9 ("CP.con" and "PT.con", respectively). A reduction by a factor of four to five in the (unweighted) average case can be observed applying the PARSETALK strategy.

Furthermore, the PARSETALK parser is able to cope with discontinuities stemming from un- or extragrammatical input anyway. The performance of a revised version of the chart parser which also handles these cases is given as "CP.disc.syn/con" in the above graphics. The missing value at sentence 10 results from the chart parser crashing on some input because of space restrictions of the run time system (the experiments were conducted with a SPARC-station 10 with 64 MB of main memory). The average reduction in comparison with this version of the chart parser is about six to nine.

Though persuasive at first sight, the superior efficiency of the PARSETALK parser is due to the incomplete depth-first nature of our approach and, thus, it is gained at the risk of not finding a correct analysis at all. A comparison of the *effectiveness* of the PARSETALK system based on the conceptual representations extracted from input texts is currently under way.